\documentclass{acsa}

\let\labelindent\relax
\usepackage{algorithm,algorithmic,amsmath,amssymb,amsthm,balance,booktabs}
\usepackage{breakurl,color,colortbl,cuted,diagbox,enumitem}
\usepackage{float,graphicx,lipsum,lscape,makecell,mathtools,microtype,mparhack}
\usepackage{multirow,outlines,pifont,rotating,tabularx,tikz,times,url,xcolor}
\usepackage{subcaption}
\usepackage[colorlinks,urlcolor=black,linkcolor=black,citecolor=black]{hyperref}

\usetikzlibrary{shapes.geometric, positioning, arrows, bayesnet}

\definecolor{myblue}{RGB}{0, 93, 170}

\jname{\textcolor{myblue}{IEEE Intelligent Systems 38(6)}}
\jmonth{November/December}
\pubyear{\the\year}

\newcommand{\ie}{i.\,e.\,, }

\newcommand{\cmark}{\ding{51}}%

\makeatletter
\newcommand{\printfnsymbol}[1]{%
  \textsuperscript{\@fnsymbol{#1}}%
}
\makeatother

\newif\iflongformat
\longformattrue

\newcommand{\furl}[1]{\footnote{\url{http://#1}}}

\newenvironment{myquote}%
  {\it \list{}{\leftmargin=0.15in\rightmargin=0.15in}\item[]}%
  {\endlist}

\newcommand{\newsection}[1]{\section{\uppercase{#1}}}

\begin{document}

\sptitle{Department: Affective Computing and Sentiment Analysis}
\editor{Editor: Erik Cambria, Nanyang Technological University}

\title{Can ChatGPT's Responses Boost Traditional Natural Language Processing?}

\author{Mostafa M. Amin}
\affil{University of Augsburg; SyncPilot GmbH}

\author{Erik Cambria}
\affil{Nanyang Technological University}

\author{Bj\"orn W. Schuller}
\affil{University of Augsburg; Imperial College London}

\begin{abstract}
The employment of foundation models is steadily expanding, especially with the launch of ChatGPT and the release of other foundation models. These models have shown the potential of emerging capabilities to solve problems, without being particularly trained to solve. A previous work demonstrated these emerging capabilities in affective computing tasks; the performance quality was similar to traditional Natural Language Processing (NLP) techniques, but falling short of specialised trained models, like fine-tuning of the RoBERTa language model.
In this work, we extend this by exploring if ChatGPT has novel knowledge that would enhance existing specialised models when they are fused together.
We achieve this by investigating the utility of verbose responses from ChatGPT about solving a downstream task, in addition to studying the utility of fusing that with existing NLP methods.
The study is conducted on three affective computing problems, namely sentiment analysis, suicide tendency detection, and big-five personality assessment.
The results conclude that ChatGPT has indeed novel knowledge that can improve existing NLP techniques by way of fusion, be it early or late fusion.
\end{abstract}

\maketitle

\label{sec:introduction}

\chapterinitial{With the recent} rapid growth of foundation models~\cite{Bommasani21-Foundation} as large language models (LLMs)~\cite{Zhau23-FoundationHistory}, a potential has appeared for emerging capabilities~\cite{Wei22-EmergLLMs} of such models to perform new downstream tasks or solve new problems, that they were not particularly trained on in the first place.

This includes models like GPT-3.5~\cite{Ouyang22-InstructGPT},
GPT-4~\cite{OpenAI23-GPT4},  
LLaMA~\cite{Touvron23-LLaMA},  
and RoBERTa~\cite{Liu19-RoBERTa}.
The capabilities of such foundation models are being explored in various domains, like
affective computing~\cite{Amin23-WAC},  
Neural Machine Translation (NMT)~\cite{Hendy23-ChatGPT-NMT}, 
agents playing games~\cite{Wang23-Voyager},  
sentiment analysis~\cite{Zhang23-SentAI},  
and general artificial intelligence~\cite{Bubeck23-SparksAGI}. 

The phenomenon of emerging capabilities of LLMs~\cite{Wei22-EmergLLMs} was more pronounced with the utilisation of fine-tuning techniques like Reinforcement Learning with Human Feedback (RLHF), as it was employed in InstructGPT~\cite{Ouyang22-InstructGPT}, which was later included in GPT-3.5 and GPT-4 models, the main underlying models of ChatGPT.

In a previous study~\cite{Amin23-WAC}, we studied the emerging capabilities of ChatGPT to solve affective computing problems, as compared to \emph{specialised} models trained on a particular problem.
The study has indeed shown the emergence of such capabilities in affective computing problems (\cite{Schuller13-CPE,Cambria2017-ACSA})  
like sentiment analysis, suicide tendency detection, and personality traits assessment.
The performance was comparable to classical Natural Language Processing (NLP) models like Word2Vec~\cite{Mikolov13-Word2Vec}, or  
Bag-of-Words (BoW)~\cite{Bishop2006}, but not better than fine-tuned LLMs like RoBERTa~\cite{Liu19-RoBERTa}.
Another issue that was encountered was parsing the results from the responses of ChatGPT, since it frequently formatted the responses differently despite being prompted to respond with a specific format.
The aforementioned conclusions had a follow up question, whether foundation models contain novel knowledge that is not acquired by specialised training of NLP models, hence leading to better results in the scenarios when fusing foundation models with specialised models. We mainly investigate this question in this study.
The contributions of this paper are as follows:
\begin{enumerate}
    \item We introduce how to prompt ChatGPT to give verbose responses that solve affective computing problems, we demonstrate this in sentiment analysis, suicide and depression detection, and big-five personality traits assessment.
    \item We present the utility of employing the verbose responses of ChatGPT when they are processed with traditional NLP techniques.
    \item We introduce how to fuse ChatGPT with existing NLP methods for affective computing, and investigate their different combinations with different fusion methods.
\end{enumerate}

The remainder of the paper is organised as follows: in the next section, we discuss related work; then, we introduce our method; afterwards, we present and discuss the results; finally, we propose concluding remarks.

\newsection{Related Work}
We focus on related work within the area of foundation models in affective-computing-related tasks (in the text domain) or hybrid formulations between foundation models and traditional NLP methods.
Both~\cite{Chen23-TransFusion,Li23-Prompt-MNER} explore a fusion between ChatGPT and other transformer-based models for Named Entity Recognition (NER).
All of~\cite{Kocon23-ChatGPT-Jack,Zhong23-ChatGPT-understanding,Qin23-chatGPT-general-solver} investigate the capabilities of ChatGPT on various NLP tasks including affective computing tasks like sentiment analysis or emotion recognition, and others like NER and text summarisation.
\cite{Zhang23-SentAI} investigates the performance of ChatGPT in several in sentiment analysis and aspect extraction.

\newsection{Method}
\label{sec:method}
Our method consists of the following components:
\begin{enumerate}
\item Prompting ChatGPT to estimate an affective answer about a given input example, thus having two texts representing a given example, namely the original text and the corresponding response of ChatGPT.
\item Process any of the two texts via traditional NLP techniques to represent them as static features vectors; we adopt RoBERTa features extracted by the RoBERTa-base LLM~\cite{Liu19-RoBERTa} or normalised BoW count vectors~\cite{Bishop2006}.
\item Train classical machine learning models on these features either by applying early fusion, by concatenating the features then training, or late fusion by training two models and averaging their prediction probabilities.
\end{enumerate}

\begin{figure*}[!t]

\centering
\scalebox{0.75}{
\begin{tikzpicture}
[node distance = 1cm, auto,font=\footnotesize,
every node/.style={node distance=3cm},
comment/.style={rectangle, inner sep= 5pt, text width=4cm, node distance=0.25cm, font=\scriptsize\sffamily},
force/.style={rectangle, draw, fill=black!10, inner sep=5pt, text width=1.3cm, text badly centered, minimum height=1cm, font=\bfseries\footnotesize\sffamily}]

\node [force] (input) {Text};
\node [force, right=7cm of input] (RoBERTa1) {RoBERTa};
\node [force, right=1cm of RoBERTa1] (mlp1) {MLP};
\node [force, right=2cm of mlp1] (label1) {Label};

\node [force, below=0.55cm of input] (input2) {Text};
\node [force, below=0.55cm of RoBERTa1] (BoW1) {BoW};
\node [force, right=1cm of BoW1] (mlp2) {MLP};
\node [force, below=0.55cm of label1] (label3) {Label};

\node [force, below=0.55cm of input2] (p1) {Text};
\node [force, right=1cm of p1] (prompt1) {Prompt};
\node [force, right=1cm of prompt1] (gpt1) {ChatGPT};
\node [force, below=0.55cm of BoW1] (RoBERTa2) {RoBERTa};
\node [force, right=1cm of RoBERTa2] (mlp3) {MLP};
\node [force, below=0.55cm of label3] (label2) {Label};

\node [force, below=0.55cm of p1] (p2) {Text};
\node [force, right=1cm of p2] (prompt2) {Prompt};
\node [force, right=1cm of prompt2] (gpt2) {ChatGPT};
\node [force, below=0.55cm of RoBERTa2] (BoW2) {BoW};
\node [force, right=1cm of BoW2] (mlp4) {MLP};
\node [force, below=0.55cm of label2] (label4) {Label};

\plate[] {plate1} {(RoBERTa1) (RoBERTa2) (BoW1) (BoW2)} {\large Early Fusion};
\plate[] {plate2} {(label1) (label2) (label3) (label4)} {\large Late Fusion};

\path[->,thick] 
(input) edge (RoBERTa1)
(RoBERTa1) edge (mlp1)
(mlp1) edge (label1)
(p1) edge (prompt1)
(prompt1) edge (gpt1)
(gpt1) edge (RoBERTa2)
(RoBERTa2) edge (mlp3)
(mlp3) edge (label2)
(input2) edge (BoW1)
(BoW1) edge (mlp2)
(mlp2) edge (label3)
(p2) edge (prompt2)
(prompt2) edge (gpt2)
(gpt2) edge (BoW2)
(BoW2) edge (mlp4)
(mlp4) edge (label4)
;

\end{tikzpicture} 
}
\caption{Pipelines of the different fusion methods. Each branch shows a single modality of selecting an input text and processing it with an NLP technique.
The input text is either used directly or by using a corresponding response from ChatGPT about it.
Subsequently, it is processed by RoBERTa or BoW.
MLPs are then used on the features to predict the binary classification labels.
We select specific branches to carry out different fusion methods.
For early fusion, the features from the selected branches are concatenated, then one MLP is used on that to predict a label.
For late fusion, the prediction scores from the single branches are averaged to give a classification probability.
}
\label{fig:pipeline}
\end{figure*}
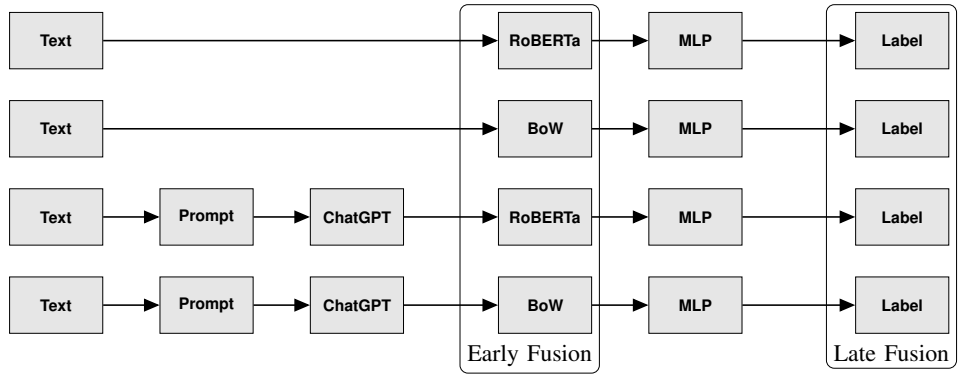

In this section, we present first the datasets for the different affective computing problems.
Afterwards, we introduce the prompting of ChatGPT, then the methods for extracting features.
Subsequently, we present how we train and tune the machine learning models.
Finally, we present a simple baseline based on ChatGPT responses.
The pipeline of our method is presented in Figure~\ref{fig:pipeline}.

\subsection{Datasets}
\label{subsec:datasets}

We present here the adopted datasets for the three affective computing problems.
A summary of their statistics is in Table~\ref{tab:data-stats}.

\begin{table}[!t]
    \centering
    \begin{tabular}{c|c||c|c|c||c|c}
    \multicolumn{2}{c|}{Dataset} & Train & Dev & Test  & +ve & -ve\\
    \hline
    \multicolumn{2}{c|}{Sentiment} & 20,000 & 5,000 & 3,000 & 1,516 & 1,484 \\
    \hline
    \multicolumn{2}{c|}{Suicide} & 9,999 & 3,881 & 2,375 & 757 & 1,618\\
    \hline
    \multirow{5}{*}{\rotatebox{90}{Personality}} &
    O &  \multirow{5}{*}{5,992} & \multirow{5}{*}{2,000} & \multirow{5}{*}{1,997} & 1,336 & 661 \\
    &C & &  &  & 1,133 & 864\\
    &E & &  &  & 890 & 1,107 \\
    &A & &  &  & 1,332 & 665\\
    &N & &  &  & 1,122 & 875\\
    \end{tabular}
    \vspace{0.2cm}
    \caption{Datasets sizes statistics, including the counts of the positive and negative classes in the Test set.
    } 
    \vspace{-0.3cm}
    \label{tab:data-stats}
\end{table}

\subsubsection{Sentiment Dataset}
We make use of the Twitter Sentiment140 dataset~\cite{Go2009-Twitter} for sentiment analysis.\footnote{We acquired the dataset from \url{https://huggingface.co/datasets/sentiment140}, on 09.02.2023.}
The dataset consists of tweets that were collected from Twitter.
Tweets are generally very noisy texts.
The dataset consists of tweets and the corresponding binary sentiment labels (positive, or negative).
The original dataset consists of 1,600,000 Tweets, however, we filtered these down into a total of 28,000 examples.
\footnote{\label{urlnote}\url{https://github.com/mostafa-mahmoud/chat-gpt-fusion-evaluation}}
We do not make use of the original Test portion in the dataset, since it consists of only 497 Tweets, and it also contains a `neutral' label unlike the rest of the dataset.
We split the original training portion into three parts as shown in Table~\ref{tab:data-stats}.

\subsubsection{Suicide and Depression Dataset}

The Suicide and Depression dataset~\cite{Desu22-Suicide} was gathered from the platform Reddit. The collection was gathered under different categories (subreddits), namely ``depression'', ``SuicideWatch'', and ``teenagers''.\footnote{We acquired the dataset on 28.01.2023 from \url{https://www.kaggle.com/datasets/nikhileswarkomati/suicide-watch}} 
The `non-sucide' label was given to the posts from the ``teenagers'' category, while the remaining texts were given the label `suicide'.
After excluding examples longer than 512 characters and downsampling the dataset, we acquired a dataset of size 16,266 that we divide into three portions Train, Dev, and Test as shown in Table~\ref{tab:data-stats}, since the original dataset was not split.

\subsubsection{Personality Dataset}
We make use of the First Impressions (FI) dataset (\cite{Ponce16-ChaLearn,Escalante17-DEML})  
for the personality task\footnote{We acquired the dataset on 03.02.2023 from \url{https://chalearnlap.cvc.uab.cat/dataset/24/description/}}. 
The big-five personality traits (OCEAN) are the traits used to represent personality, namely,
\emph{Openness to experience}, \emph{Conscientiousness}, \emph{Extraversion}, \emph{Agreeableness}, and \emph{Neuroticism}.
The dataset was gathered by collecting videos from YouTube, and slicing them into 15 seconds clips with one speaker.
The personality labels were collected through crowdsourcing using Amazon Mechanical Turk (AMT), by making pair comparisons between different videos.
Each personality trait is represented by a continuous regression value within $[0, 1]$.
In our setup, we utilise only the text modality of the entire FI dataset, with its provided split, originating from the transcriptions of the videos.
We train regression models (by employing Mean Absolute Error as a loss function)~\cite{Kaya17-PersSOTA},
because the continuous labels give a granular estimation of the personality labels.
For evaluation, we interpret the predicted regression labels as the probability of the positive class, which is equivalent to binarising the labels with the threshold $0.5$.

\subsection{ChatGPT Prompts}
\label{subsec:prompts}

To formulate the ChatGPT text modalities, we need to formulate a prompt to ask ChatGPT, in order to obtain a reasonable answer.
We formulate a prompt for each specific problem to ask it about the label.
First, we design the prompt to ask for a binary label of the corresponding problem, while emphasising narrowing down the answer to only two labels while excluding more `neutral' labels. 
Similar to a previous work~\cite{Amin23-WAC}, we design the prompts to have the disclaimer \emph{It does not have to be fully correct}, and ask \emph{what is your guess for the answer}, instead of \emph{What is the answer} or \emph{Can you guess the answer}.
This formulation is to avoid ChatGPT from responding that it is not sure about the answer, hence not giving any answer.
Unlike~\cite{Amin23-WAC}, we ask ChatGPT to be verbose and explain the reasoning behind the answer, since we are processing that with NLP methods (unlike~\cite{Amin23-WAC}, where the final label was parsed).
A last sentence is added to avoid redundant disclaimer in the response of ChatGPT.

We make use of the OpenAI API to use ChatGPT\footnote{\url{https://platform.openai.com/docs/guides/gpt/chat-completions-api}}, using the the model `gpt-3.5-turbo-0301'.
We do not give a system message, we just use the prompt corresponding to the specific problem as the only user message in the input conversation, with the input text of the example.
The assistant response is what we use as the response of ChatGPT.
We use the default parameters for generation, namely the answer with highest score ($n=1$), and the temperature parameter $T=1.0$.

The prompts for the given problems are given below, by substituting the input \emph{\{text\}}.
For the personality traits, we query the API five times for each of the five traits by substituting the \emph{\{trait\}}.
\\

\begin{itemize}[wide, labelindent=0pt,topsep=0pt,noitemsep,leftmargin=4pt]

    \item The prompt for the sentiment classification:
\begin{myquote}
What is your guess for the sentiment of the text ``\{text\}''?
Answer positive or negative, but not neutral.
Try to narrow down the answer to be one of those two.
It does not have to be fully correct.
Explain your answer briefly. 
Do not show any warning after.
\end{myquote}

    \item The prompt for the suicide detection:
\begin{myquote}
What is your guess, is a person saying the text ``\{text\}'' has suicide tendencies?
Answer yes or no.
It does not have to be fully correct.
Explain your answer briefly.
Do not show any warning after.
\end{myquote}

    \item The prompt for the personality traits:
\begin{myquote}
What is your guess for the personality trait ``\{trait\}'', 
from the big-five personality traits, 
of someone who said ``\{text\}''?
Answer low or high, but not neutral. 
Try to narrow down the answer to low or high. 
It does not have to be fully correct. 
Explain your answer briefly. 
Do not show any warning after.

\end{myquote}

\end{itemize}

Regarding the number of tokens, a token on average gives 4.3 characters.
There is an overhead of 8 tokens that gets added for each call to the API.
The prompts using an empty input string for \emph{`\{text\}'} consist of 
63, 50, 75 tokens, for the sentiment, suicide, and personality prompts, respectively.
Processing a prompt of total $T$ tokens (system, prompted input and output) took an average of $0.038 T + 1.32$\,sec.

\subsection{Text Features}
\label{subsec:features}

In order to process the text, we need to extract features from it.
We employ two ways to extract features, one via the LLM RoBERTa~\cite{Liu19-RoBERTa}, and $n$-gram BoW.

\subsubsection{RoBERTa Language Model}
The RoBERTa~\cite{Liu19-RoBERTa} feature set is obtained by the pretrained LLM RoBERTa, which is based on the BERT model with a transformer architecture.
The model has two variants; we utilise the smaller variant, namely RoBERTa-base\footnote{Acquired on 09.02.2023 from \url{https://huggingface.co/docs/transformers/model_doc/roberta}}.
The model was trained on large datasets with reddit posts and English Wikipedia, and English news~\cite{Liu19-RoBERTa}.
In order to extract the embedding for a string, it is first encoded with a subword encoder then fed to the RoBERTa model to give a sequential set of features with attention weights.
These are reduced through a pooling layer in the model to produce the final static vector of 768 features representing the given string.

\subsubsection{Bag of Words}
The BoW feature set is achieved by constructing $n$-grams,
and then using the classical term-frequency inverse-document-frequency (TF-IDF) to count each term while normalising them by the frequency across all documents~\cite{Bishop2006}.
For the input texts, we keep only the most common 10,000 words (\ie $1$-grams), to give a static vector of 10,000 features representing the text.
For the responses of ChatGPT, we utilise the most common 2,000 $n$-grams ($n \in\{1,2,3\}$).
The vectors are scaled by the maximum absolute values to be within the range $[-1,1]$.
The reason we utilise $n$-grams for ChatGPT responses is that, it is common that ChatGPT would give prediction expressions like `high extraversion', or `sentiment is negative'.

\subsection{Models and tuning}
\label{subsec:models}
Given a feature set (or a fusion thereof) we train a Multi-Layer Perceptron (MLP)~\cite{Bishop2006} to predict the final label.
We opt to use MLPs, because preliminary experiments showed that MLPs were performing slightly better than Support Vector Machines (SVMs)~\cite{Bishop2006} for the given tasks.
We construct an MLP with $N$ hidden layers, with $U$ units in the first hidden layer, then each following hidden layer has half the number of neurons of the hidden layer preceding it (we cap this number to be at least 32 units).
ReLU is the activation function used for all layers, except for the final layer, where we apply sigmoid to predict the final label within the range $[0,1]$.
We leverage Adam~\cite{Diederik15-Adam} as an optimisation algorithm, with a learning rate $\alpha$.
The loss function is either Mean Absolute Error (MAE) for regression training (for personality training), or otherwise negative log likelihood for classification training.
We employ the hyperparameter optimisation toolkit SMAC~\cite{Lindauer22-SMAC} to select the best hyperparameters for each problem/dataset and each input modality (or early fusion combinations thereof).
We explore 20 hyperparameters samples for each problem.
The hyperparameter space has $N \in [0, 3]$, $U \in [64, 512]$ (log-sampled), and $\alpha \in [10^{-6}, 10]$ (log-sampled).

\subsection{Fusion}
We deploy early fusion by concatenating the features extracted by RoBERTa or BoW, then training one MLP on the concatenated vector similar to training a single method.
On the other hand, the late fusion is achieved by averaging the probabilities predicted by the given methods.

\subsection{Baseline}

We employ a simple baseline based on the responses of ChatGPT.
In the prompts, we instruct ChatGPT to give a binary label before explaining the answer, hence we construct the baseline to predict a label only if the word corresponding to its class is present in the text.
For sentiment analysis, the baseline would predict `positive' only if the response of ChatGPT contains the word `positive', and it would predict `negative' only if the response contains the word `negative'.
For suicide detection, the two classification keywords are `yes' and `no'.
For personality, the two keywords become `high' and `low'.
We exclude the evaluation of responses that include both words or neither, which is roughly only 5\% of the Test sets in our experiments.
The intuition behind this baseline is that it is similar to parsing the labels from the non-verbose response.

\newsection{Results}
\label{sec:experiments}

\begin{table*}[!t]
    \centering
    \begin{subtable}{\textwidth}
    \begin{tabular}{c|c|c|c||l||c||c||c||c|c|c|c|c}
    \multicolumn{2}{c|}{Text} & \multicolumn{2}{c||}{ChatGPT} & \multirow{2}{*}{Fusion} & 
    \multirow{2}{*}{Sent.} & \multirow{2}{*}{Suic.} & \multicolumn{6}{c}{Personality} \\ \cline{1-4} \cline{8-13}
    RoBERTa & BoW & RoBERTa & BoW & & & & Average & O & C & E & A & N  \\ \hline
        \multicolumn{4}{c||}{Baseline}  &    --                &      77.68  &  94.48     &   54.34  &  65.54  &  58.43  &  47.89  &  53.35  &  46.47    \\ \hline
      \cmark  &   &   &    &    --                             &      77.83  &  \textbf{95.37}     &  \underline{\textbf{64.12}}  &  \textbf{67.55}  &  \textbf{63.09}  &  \underline{\textbf{61.19}}  &  \underline{\textbf{67.55}}  &  \textbf{61.19}    \\   \cline{1-4}
        & \cmark  &   &    &    --                             &      73.90  &  90.40     &  61.66  &  66.80  &  59.89  &  57.34  &  66.80  &  57.49    \\   \cline{1-4}
        &   &  \cmark  &    &    --                            &      \textbf{80.27}  &  92.34     &  61.01  &  66.90  &  59.09  &  55.43  &  66.70  &  56.94    \\   \cline{1-4}
        &   &   &  \cmark   &    --                            &      79.83  &  91.92     &  60.71  &  66.90  &  57.24  &  55.73  &  66.70  &  56.99    \\   \hline
       \cmark &   & \cmark  &    & Early                     &      \textbf{81.20}  &  \underline{\textbf{96.17}}     &  \textbf{63.65}  &  \underline{\textbf{68.15}}  &  \textbf{61.84}  &  \textbf{60.54}  &  66.70  &  \textbf{60.99}    \\    \cline{1-4}
        & \cmark  &   & \cmark   & Early                     &      80.90  &  93.52     &  61.79  &  66.90  &  60.39  &  56.94  &  66.60  &  58.14    \\   \cline{1-4}
      \cmark  & \cmark  &   &    & Early                     &      76.27  &  92.97     &  62.21  &  67.40  &  59.69  &  59.39  &  66.60  &  57.99    \\   \cline{1-4}
        &   & \cmark  & \cmark   & Early                     &      80.03  &  91.96     &  60.89  &  66.90  &  58.29  &  55.53  &  66.60  &  57.14    \\   \cline{1-4}
      \cmark  & \cmark  &  \cmark  &  \cmark  & Early        &      80.93  &  93.94     &  61.53  &  67.00  &  60.34  &  57.04  &  \textbf{66.80}  &  56.48    \\    \hline
      \cmark  &     &  \cmark  &    & Late                   &      81.60  &  \textbf{96.13}     &  63.26  &  \textbf{66.95}  &  61.19  &  59.49  &  66.70  &  \underline{\textbf{61.99}}    \\   \cline{1-4}
        &  \cmark  &    & \cmark  & Late                     &      80.77  &  93.94     &  61.68  &  66.90  &  59.94  &  58.54  &  66.65  &  56.38    \\   \cline{1-4}
        \cmark &  \cmark  &   &    & Late                    &      79.40  &  95.54     &  \textbf{63.59}  &  66.75  &  \underline{\textbf{63.40}}  &  \textbf{60.79}  &  \textbf{66.75}  &  60.24    \\   \cline{1-4}
        &   & \cmark  & \cmark   & Late                      &      81.13  &  92.76     &  61.08  &  66.90  &  59.64  &  55.38  &  66.65  &  56.84    \\   \cline{1-4}
   \cmark     &  \cmark &  \cmark  &  \cmark  & Late         &      \underline{\textbf{82.60}}  &  95.45     &  62.66  &  66.90  &  61.49  &  59.39  &  66.70  &  58.84    \\
    
    \end{tabular}
    
    \caption{Classification accuracy results for all the problems.}
    \vspace{0.1cm}
    \end{subtable}

    \begin{subtable}{\textwidth}
    \centering
    \begin{tabular}{c|c|c|c||l||c||c||c||c|c|c|c|c}
    \multicolumn{2}{c|}{Text} & \multicolumn{2}{c||}{ChatGPT} & \multirow{2}{*}{Fusion} & 
    \multirow{2}{*}{Sent.} & \multirow{2}{*}{Suic.} & \multicolumn{6}{c}{Personality} \\ \cline{1-4} \cline{8-13}
    RoBERTa & BoW & RoBERTa & BoW & & & & Average & O & C & E & A & N  \\ \hline

        \multicolumn{4}{c||}{Baseline} &    --                 &      77.44  &  90.56     &   51.79   &  52.08  &  57.43  &  51.43  &  52.59  &  45.41    \\ \hline
      \cmark  &   &   &    &    --                             &    77.82     &  \textbf{94.07}      &  \underline{\textbf{56.76}}  &  \textbf{52.02}  &  \underline{\textbf{61.71}}  &  \underline{\textbf{58.66}}  &  \textbf{51.62}  &  \underline{\textbf{59.79}}    \\   \cline{1-4}
        & \cmark  &   &    &    --                             &    73.91     &  88.46      &  52.77  &  50.00  &  56.37  &  55.50  &  50.30  &  51.69    \\   \cline{1-4}
        &   &  \cmark  &    &    --                            &    \textbf{80.25}     &  90.37      &  51.14  &  50.04  &  53.37  &  50.00  &  50.00  &  52.29    \\   \cline{1-4}
        &   &   &  \cmark   &    --                            &    79.81     &  90.10      &  51.64  &  50.00  &  54.59  &  51.41  &  50.04  &  52.15    \\   \hline
       \cmark &   & \cmark  &    & Early                     &    \textbf{81.19}     &  \underline{\textbf{95.89}}      &  \textbf{55.90}  &  \underline{\textbf{53.65}}  &  \textbf{59.69}  &  \textbf{57.97}  &  50.04  &  \textbf{58.15}    \\    \cline{1-4}
        & \cmark  &   & \cmark   & Early                     &    80.89     &  92.04      &  53.56  &  50.08  &  57.91  &  54.37  &  50.72  &  54.73    \\   \cline{1-4}
      \cmark  & \cmark  &   &    & Early                     &    76.27     &  91.54      &  54.58  &  51.44  &  57.48  &  57.04  &  \underline{\textbf{52.67}}  &  54.28    \\   \cline{1-4}
        &   & \cmark  & \cmark   & Early                     &    80.02     &  89.99      &  51.77  &  50.00  &  55.16  &  50.44  &  49.92  &  53.30    \\   \cline{1-4}
      \cmark  & \cmark  &  \cmark  &  \cmark  & Early        &    80.91     &  92.77      &  53.13  &  50.69  &  58.06  &  53.97  &  50.34  &  52.62    \\    \hline
      \cmark  &     &  \cmark  &    & Late                   &    81.59     &  \textbf{95.12}      &  54.41  &  \textbf{50.08}  &  57.50  &  55.53  &  50.00  &  \textbf{58.97}    \\   \cline{1-4}
        &  \cmark  &    & \cmark  & Late                     &    80.76     &  92.25      &  52.40  &  50.00  &  56.08  &  55.44  &  49.96  &  50.51    \\   \cline{1-4}
        \cmark &  \cmark  &   &    & Late                    &    79.40     &  94.37      &  \textbf{55.14}  &  49.93  &  \textbf{60.75}  &  \textbf{58.25}  &  \textbf{50.19}  &  56.60    \\   \cline{1-4}
        &   & \cmark  & \cmark   & Late                      &    81.11     &  90.78      &  51.43  &  50.00  &  55.43  &  49.98  &  49.96  &  51.80    \\   \cline{1-4}
   \cmark     &  \cmark &  \cmark  &  \cmark  & Late         &    \underline{\textbf{82.59}}     &  94.20      &  53.35  &  50.00  &  57.39  &  55.42  &  50.00  &  53.92    \\

    \end{tabular}
    \caption{Unweighted Average Recall (UAR) results on all problems.}
    \vspace{0.1cm}
    \end{subtable}
    \caption{
    Results for all the problems with the different fusion methods.
    There are two text-based inputs, the original text (Text), or the verbose response of ChatGPT on a question about the original text and the corresponding problem (ChatGPT).
    Each text input is processed in two ways, using RoBERTa features or BoW.
    The features are processed with an MLP to give the final binary classification label of the problem.
    The fusion is either done on the feature level with one MLP (Early), or on the predictions level (Late).
    Marked in bold are the best results for each combination of problem and fusion.
    Underlined are the best results for each problem.
    }
    \label{tab:performance}
\end{table*}

We experiment the combinations of three main parameters, the \emph{text} to be used, the corresponding extracted \emph{features} to represent the text, and \emph{how to fuse} them.
The texts are either the original text or the corresponding response from ChatGPT.
The features are either embeddings obtained by RoBERTa, or using normalised count vectors constructed by a simple $n$-gram BoW approach.
The fusion of the models is either done (early) on the feature level, or (late) on the prediction level by averaging the probabilities of the classes.
We also include the baseline results.
The main results of the experiments are shown in Table \ref{tab:performance}, where we evaluate classification accuracy and Unweighted Average Recall (UAR), which is the unweighted average of the accuracy of classifying each class separately~\cite{Schuller13-TI2}. 
Finally, we refer to the combination of input text (original input or ChatGPT response thereof) and NLP processing technique as a \emph{modality}.

\subsection{Discussion}

First of all, both metrics show a wide agreement on the relative performance of a specific model on a specific problem, \ie the relative order of models on a specific problem based is roughly similar for both metrics.
The results of utilisng the original text (for each of the single modalities Text+RoBERTa and Text+BoW) are close to previous work~\cite{Amin23-WAC}, with a slight difference due to the different sampling from the original datasets.
The results of the single modality ChatGPT+RoBERTa are 
decent, 
comparable to the single modality Text+BoW, but worse than Text+RoBERTa in most cases except for sentiment analysis.
The results of ChatGPT+BoW are slightly worse than ChatGPT+RoBERTa.
In a similar fashion, these results of ChatGPT are resembling the previous work~\cite{Amin23-WAC}, where ChatGPT was comparable to the Text+BoW modality.
Furthermore, the aggregate performances across problems is also similar to~\cite{Amin23-WAC}, where ChatGPT was the most superior in sentiment analysis, whilst most inferior in personality assessment.

The results of fusion are inclined to show that the most competent fusion combination is adopting only Text+RoBERTa and ChatGPT+RoBERTa, whether in early or late fusion; however, the early fusion of these two modalities is showing the most superior performance in most scenarios, except the sentiment analysis.
Disregarding the specific combination of these two modalities, late fusion is performing better compared to the corresponding instances of early fusion in most cases of the other modality combinations.
For instance, the late fusion of all modalities is better than their early fusion; similarly for the combination of Text+RoBERTa and Text+BoW.

Consequently, the impact of fusion overall is not very straight forward to explain, because the \emph{single} modality Text+RoBERTa is the best for the personality assessment, while the \emph{early} fusion of Text+RoBERTa and ChatGPT+RoBERTa is the best for suicide detection, and the \emph{late} fusion of all modalities is the best for sentiment analysis.
The reason for the superiority of the single modality in the personality assessment is probably due to the poor performance of ChatGPT on the given text, since ChatGPT single modalities are the worst ones.
On the other hand, if ChatGPT has a decent performance, then applying fusion has definitely a strong improvement impact, be it early or late fusion.
However, the superiority of late fusion against early fusion depends primarily on the problem and the data distribution.

From the practical advantages of early fusion, it needs hyperparameter tuning only once, compared to the late fusion which needs to tune a model for each modality.
On the other hand, the late fusion has an architectural advantage that it can deploy different training sizes for each modality.
For instance, it is possible to train Text+RoBERTa with much larger dataset size, while training ChatGPT+RoBERTa on a smaller dataset size;
we will evaluate this in future work.

In the previous work~\cite{Amin23-WAC}, the ChatGPT results were labels that were parsed from the non-verbose responses (typically, a binary label like `low' or `high', with some variance in the formatting), whereas in this work we process the verbose response by applying NLP methods.
The effectiveness of employing the verbose responses is demonstrated by the baseline approach, where the results of the single ChatGPT modalities are close to the baseline.
The verbose responses (compared to the non-verbose ChatGPT baseline) lead to better responses for both sentiment analysis and personality assessment, but with some drop in suicide detection.
The verbose responses have the additional advantage of avoiding the problem of parsing the label from the response of ChatGPT, since the responses (including the non-verbose) do not always follow the same format despite being prompted to~\cite{Amin23-WAC}.
The last obvious advantage of verbose responses is the ability to include them in fusion models in various ways, which can lead to a much better performance as discussed earlier.

In summary, utilising the verbose responses of ChatGPT adds unique information that can yield improvements to the results of existing NLP models.
Employing fusion techniques on top of that, whether early or late fusion, will yield bigger improvements.
Moreover, it is sufficient in most cases to process the texts only with RoBERTa to extract features, for both the original text and the verbose response of ChatGPT without worrying about parsing the corresponding labels.

\newsection{Conclusion}
\label{sec:conclusion}

In this work, we explored the fusion capabilities of ChatGPT with traditional Natural Language Processing (NLP) models in affective computing problems.
We first prompted ChatGPT to give verbose responses to answer binary classification questions for three affective computing downstream tasks, namely sentiment analysis, suicide tendency detection, and big-five personality traits assessment.
Additionally, we processed the input texts and the corresponding ChatGPT responses with two NLP techniques, namely fine-tuning RoBERTa language model and $n$-gram BoW;
these features were trained by leveraging Multi-Layer Perceptrons (MLPs).
Furthermore, we investigated two fusion methods, early fusion (on the features level) or late fusion (on the prediction level).

The experiments have demonstrated that leveraging ChatGPT verbose responses bears novel knowledge in affective computing 
and probably beyond, which should be evaluated next,
that can aid existing NLP techniques by ways of fusion, whether early or late fusion.
First, we demonstrated the benefit of using verbose responses while processing them with NLP techniques, as compared to parsing classification labels from the non-verbose labels.
Subsequently, this provided the possibility of seamlessly fusing ChatGPT responses with existing NLP methods, hence achieving a better performance via both early or late fusions.
Furthermore, the experiments have demonstrated that utilising only RoBERTa to process and fuse the input texts and ChatGPT responses (with an inclination to early fusion than late) can be sufficient to reach the best performance.

\bibliographystyle{acsa}
\bibliography{references}

\begin{thebibliography}{10}
\newcommand{\enquote}[1]{``#1''}
\providecommand{\url}[1]{\texttt{#1}}
\providecommand{\urlprefix}{}
\expandafter\ifx\csname urlstyle\endcsname\relax
  \providecommand{\doi}[1]{doi:\discretionary{}{}{}#1}\else
  \providecommand{\doi}{doi:\discretionary{}{}{}\begingroup
  \urlstyle{rm}\Url}\fi

\bibitem{Bommasani21-Foundation}
R.~Bommasani et~al., \enquote{{On the Opportunities and Risks of Foundation
  Models},} \emph{arXiv:2108.07258}, 2021.

\bibitem{Zhau23-FoundationHistory}
C.~Zhou et~al., \enquote{{A Comprehensive Survey on Pretrained Foundation
  Models: A History from BERT to ChatGPT},} \emph{arXiv:2302.09419}, 2023.

\bibitem{Wei22-EmergLLMs}
J.~Wei et~al., \enquote{{Emergent Abilities of Large Language Models},}
  \emph{arXiv:2206.07682}, 2022.

\bibitem{Ouyang22-InstructGPT}
L.~Ouyang et~al., \enquote{{Training language models to follow instructions
  with human feedback},} \emph{arXiv:2203.02155}, 2022.

\bibitem{OpenAI23-GPT4}
OpenAI, \enquote{{GPT-4 Technical Report},} \emph{arXiv:2303.08774}, 2023.

\bibitem{Touvron23-LLaMA}
H.~Touvron et~al., \enquote{{LLaMA: Open and Efficient Foundation Language
  Models},} \emph{arXiv:2302.13971}, 2023.

\bibitem{Liu19-RoBERTa}
Y.~Liu et~al., \enquote{{RoBERTa: A Robustly Optimized BERT Pretraining
  Approach},} \emph{arXiv:1907.11692}, 2019.

\bibitem{Amin23-WAC}
M.~M. Amin, E.~Cambria, and B.~W. Schuller, \enquote{{Will Affective Computing
  Emerge from Foundation Models and General AI? A First Evaluation on
  ChatGPT},} \emph{IEEE Intelligent Systems}, vol.~38, no.~2, 2023, pp. 15--23.

\bibitem{Hendy23-ChatGPT-NMT}
A.~Hendy et~al., \enquote{{How Good Are GPT Models at Machine Translation? A
  Comprehensive Evaluation},} \emph{arXiv:2302.09210}, 2023.

\bibitem{Wang23-Voyager}
G.~Wang et~al., \enquote{{Voyager: An Open-Ended Embodied Agent with Large
  Language Models},} \emph{arXiv:2305.16291}, 2023.

\bibitem{Zhang23-SentAI}
W.~Zhang et~al., \enquote{{Sentiment Analysis in the Era of Large Language
  Models: A Reality Check},} \emph{arXiv:2305.15005}, 2023.

\bibitem{Bubeck23-SparksAGI}
S.~Bubeck et~al., \enquote{{Sparks of Artificial General Intelligence: Early
  experiments with GPT-4},} \emph{arXiv:2303.12712}, 2023.

\bibitem{Schuller13-CPE}
B.~Schuller and A.~Batliner, \emph{{Computational Paralinguistics: Emotion,
  Affect and Personality in Speech and Language Processing}}, John Wiley \&
  Sons, New York City, NY, USA, 2013.

\bibitem{Cambria2017-ACSA}
E.~Cambria, \enquote{{Affective Computing and Sentiment Analysis},} \emph{IEEE
  Intelligent Systems}, vol.~31, no.~2, 2016, pp. 102--107.

\bibitem{Mikolov13-Word2Vec}
T.~Mikolov et~al., \enquote{{Distributed Representations of Words and Phrases
  and their Compositionality},} \emph{Advances in Neural Information Processing
  Systems}, Curran Associates, Inc., Lake Tahoe, NV, USA, 2013.

\bibitem{Bishop2006}
C.~M. Bishop, \emph{{Pattern Recognition and Machine Learning}}, Springer, New
  York City, NY, USA, 2006.

\bibitem{Chen23-TransFusion}
Y.~Chen, V.~Shah, and A.~Ritter, \enquote{{Better Low-Resource Entity
  Recognition Through Translation and Annotation Fusion},}
  \emph{arXiv:2305.13582}, 2023.

\bibitem{Li23-Prompt-MNER}
J.~Li et~al., \enquote{{Prompt ChatGPT In MNER: Improved multimodal named
  entity recognition method based on auxiliary refining knowledge from
  ChatGPT},} \emph{arXiv:2305.12212}, 2023.

\bibitem{Kocon23-ChatGPT-Jack}
J.~Koco{\'n} et~al., \enquote{{ChatGPT: Jack of all trades, master of none},}
  \emph{Information Fusion}, vol.~99, 2023, p. 101861.

\bibitem{Zhong23-ChatGPT-understanding}
Q.~Zhong et~al., \enquote{{Can ChatGPT Understand Too? A Comparative Study on
  ChatGPT and Fine-tuned BERT},} \emph{arXiv:2302.10198}, 2023.

\bibitem{Qin23-chatGPT-general-solver}
C.~Qin et~al., \enquote{{Is ChatGPT a General-Purpose Natural Language
  Processing Task Solver?}} \emph{arXiv:2302.06476}, 2023.

\bibitem{Go2009-Twitter}
A.~Go, R.~Bhayani, and L.~Huang, \enquote{{Twitter Sentiment Classification
  using Distant Supervision},} \emph{CS224N project report, Stanford}, 2009, p.
  2009.

\bibitem{Desu22-Suicide}
V.~Desu et~al., \enquote{{Suicide and Depression Detection in Social Media
  Forums},} \emph{Smart Intelligent Computing and Applications, Volume 2},
  Springer Nature Singapore, Singapore, Singapore, 2022, pp. 263--270.

\bibitem{Ponce16-ChaLearn}
V.~Ponce-L\'opez et~al., \enquote{{Chalearn lap 2016: First Round Challenge on
  First Impressions - Dataset and Results},} \emph{European conference on
  computer vision}, Springer International Publishing, Cham, Switzerland, 2016,
  pp. 400--418.

\bibitem{Escalante17-DEML}
H.~J. Escalante et~al., \enquote{{Design of an Explainable Machine Learning
  Challenge for Video Interviews},} \emph{International Joint Conference on
  Neural Networks (IJCNN)}, IEEE, Anchorage, AK, USA, 2017, pp. 3688--3695.

\bibitem{Kaya17-PersSOTA}
H.~Kaya, F.~Gurpinar, and A.~Ali~Salah, \enquote{{Multi-Modal Score Fusion and
  Decision Trees for Explainable Automatic Job Candidate Screening From Video
  CVs},} \emph{Conference on Computer Vision and Pattern Recognition
  Workshops}, IEEE, Honolulu, HI, USA, 2017, pp. 1--9.

\bibitem{Diederik15-Adam}
D.~P. Kingma and J.~Ba, \enquote{{Adam: A Method for Stochastic Optimization},}
  \emph{3rd International Conference on Learning Representations, Conference
  Track Proceedings}, ICLR, San Diego, CA, USA, 2015.

\bibitem{Lindauer22-SMAC}
M.~Lindauer et~al., \enquote{{SMAC3: A Versatile Bayesian Optimization Package
  for Hyperparameter Optimization},} \emph{Journal of Machine Learning
  Research}, vol.~23, no.~54, 2022, pp. 1--9.

\bibitem{Schuller13-TI2}
B.~Schuller et~al., \enquote{{The INTERSPEECH 2013 Computational
  Paralinguistics Challenge: Social Signals, Conflict, Emotion, Autism},}
  \emph{Proceedings INTERSPEECH}, ISCA, Lyon, France, 2013, pp. 148--152.

\end{thebibliography}

\begin{IEEEbiography}{Mostafa M. Amin}{\,}is currently working toward the Ph.D.~degree with the Chair of Embedded Intelligence for Health Care and Wellbeing with University of Augsburg, while working as Senior Research Data Scientist at SyncPilot GmbH in Augsburg, Germany. His research interests include Affective Computing, Audio and Text Analytics. He received a M.Sc.\ degree in Computer Science from the University of Freiburg, Germany. Contact him at \href{mailto:first.author@institution.edu}{mostafa.mohamed@uni-a.de} 
\end{IEEEbiography}

\begin{IEEEbiography}{Erik Cambria}{\,}is a professor of Computer Science and Engineering at Nanyang Technological University, Singapore. His research focuses on neurosymbolic AI for explainable natural language processing in domains like sentiment analysis, dialogue systems, and financial forecasting. He is an IEEE Fellow and a recipient of several awards, e.\,g., IEEE Outstanding Career Award, was listed among the AI's 10 to Watch, and was featured in Forbes as one of the 5 People Building Our AI Future. Contact him at \href{mailto:cambria@ntu.edu.sg}{cambria@ntu.edu.sg}.
\end{IEEEbiography}

\begin{IEEEbiography}{Bj\"orn W. Schuller}{\,}is currently a professor of Artificial Intelligence with the Department of Computing, Imperial College London, UK, where he heads the Group on Language, Audio, \& Music (GLAM). He is also a full professor and the head of the Chair of Embedded Intelligence for Health Care and Wellbeing with the University of Augsburg, Germany, and the Founding CEO/CSO of audEERING.  He is an IEEE Fellow alongside other Fellowships. 
Contact him at \href{mailto:cambria@ntu.edu.sg}{schuller@IEEE.org}.
\end{IEEEbiography}

\end{document}